\documentclass[lettersize,journal]{IEEEtran}
\usepackage{amsmath,amsfonts}
\usepackage{algorithmic}
\usepackage{algorithm}
\usepackage{array}
\usepackage[caption=false,font=normalsize,labelfont=sf,textfont=sf]{subfig}
\usepackage{textcomp}
\usepackage{stfloats}
\usepackage{url}
\usepackage{verbatim}
\usepackage{graphicx}
\usepackage{cite}
\usepackage{float}

\usepackage{booktabs} 
\usepackage{threeparttable}
\usepackage{multirow} 
\usepackage{enumitem}
\usepackage{colortbl}
\usepackage[table,dvipsnames]{xcolor}  

\usepackage{caption}
\usepackage{lscape}

\definecolor{LightGray}{gray}{0.90}

\makeatletter
\renewcommand\p@subsubsection{} 
\makeatother
\usepackage[colorlinks=true,linkcolor=blue,citecolor=blue,urlcolor=black]{hyperref}
\usepackage{cleveref}

\begin{document}

\title{Style-Adaptive Detection Transformer for Single-Source Domain Generalized Object Detection}

\author{ 
Jianhong Han,~\IEEEmembership{Student Member,~IEEE},
Yupei Wang$^{*}$,~\IEEEmembership{Member,~IEEE},
Liang Chen,~\IEEEmembership{Member,~IEEE} 

\IEEEcompsocitemizethanks{
\IEEEcompsocthanksitem This work was supported by National Natural Science Foundation of China under Grant 62301046, National Key Laboratory for Space-Born Intelligent Information Processing under Grant TJ-01-22-01. \textit{
(Corresponding author: Yupei Wang.)}
\IEEEcompsocthanksitem J. Han, Y. Wang and L. Chen  are with the School of Information and Electronics, Beijing Institute of Technology, Beijing 100081, China, also with the Beijing Institute of Technology Chongqing Innovation Center, Chongqing 401135, China, and also with the National Key Laboratory for Space-Born Intelligent Information Processing, Beijing 100081, China. E-mail: hanjianhong1996@163.com, wangyupei2019@outlook.com, chenl@bit.edu.cn. 
}
\thanks{Manuscript submitted to IEEE Transactions on Circuits and Systems for Video Technology.}
}



\maketitle

\begin{abstract}
Single-source Domain Generalization (SDG) in object detection aims to develop a detector using only data from source domain that can exhibit strong generalization capability when applied to other unseen target domains. Existing methods are built upon CNN-based detectors and primarily improve their robustness by employing carefully designed data augmentation strategies integrated with feature alignment techniques. However, the data augmentation methods have inherent drawbacks that are only effective when the augmented sample distribution approximates or covers the unseen scenarios, thus failing to enhance the detector's generalization capability across all unseen scenarios. Furthermore, while the recent DEtection TRansformer (DETR) has demonstrated superior generalization capability in domain adaptation tasks due to its efficient global information extraction, its potential for SDG tasks remains unexplored. To this end, we introduce a strong DETR-based detector named the Style-Adaptive DEtection TRansformer (SA-DETR) for SDG in object detection. Specifically, we present an online domain style adapter to project the style representation of the unseen target domain into the training domain. This adapter maintains a dynamic memory bank that self-organizes onto diverse style bases and continuously absorbs information from unseen domains within a test-time adaptation framework, ultimately delivering precise style adaptation. Then, the object-aware contrastive learning module is proposed to force the detector to extract domain-invariant features of objects through contrastive learning. By employing carefully designed object-aware gating masks to constrain the scope of feature aggregation in both spatial position and semantic category, this module effectively achieves cross-domain contrast of instance-level features, thereby further enhancing the detector’s generalization capability. Extensive experimental results demonstrate the superior performance and generalization capability of our proposed SA-DETR across five different weather scenarios.  Code is released at \href{https://github.com/h751410234/SA-DETR}{https://github.com/h751410234/SA-DETR}.

\end{abstract}


\section{Introduction}
\IEEEPARstart{T}{he} task of Single-source Domain Generalization (SDG) is designed to mitigate domain gaps that arise from distribution differences between the training data (source domain) and real-world application data (target domain). Unlike traditional unsupervised domain adaptation tasks\cite{Chen_Li_Sakaridis_Dai_Van_Gool_2018,zhang2023detr,Yu_Liu_Wei_Zhou_Nakata_Gudovskiy_Okuno_Li_Keutzer_Zhang_2022}, which allow the use of images from the target domain, SDG enhances the robustness of networks to various unseen target scenarios by using only data from a single-source domain. Although this task is more difficult, it has been receiving increasing attention because it better aligns with practical deployment requirements, such as autonomous driving.

Recently, numerous single-source domain generalization (SDG) methods~\cite{wu2022single,vidit2023clip,liu2024unbiased} for object detection have been proposed, employing carefully designed generalization strategies to enhance detector robustness, yet they still suffer from two major limitations. First, to overcome the limited sample space of single-domain data, existing methods typically employ innovative data augmentation strategies (at the image or feature level) to obtain more diverse samples for training, thereby enhancing the generalization capability of detectors for unseen scenarios. These methods have inherent drawbacks that are only effective when the augmented sample distribution approximates or covers the unseen scenarios. Due to the unpredictable deployment environments, which are likely to have significant differences from the augmented samples, the improvement in detector performance is minimal in such cases. Second, another line of research tackles this issue through style modulation, suppressing domain‑specific styles via instance normalization~\cite{Pan_Luo_Shi_Tang_2018} and feature whitening~\cite{Choi_Jung_Yun_Kim_Kim_Choo_2021}, or rectifying input images with simply constructed style bases~\cite{11016953,11023626}. Such a static treatment overlooks the high diversity and dynamic nature of styles in the wild. Consequently, it remains insufficiently adaptive when confronted with large domain gaps.

Moreover, researchers have primarily explored and demonstrated effectiveness within the Faster R-CNN\cite{Ren_He_Girshick_Sun_2017}, while transformer-based detectors have been less studied. Recent works\cite{Wang_Cao_Zhang_He_Zha_Wen_Tao_2021,Huang_Lu_Lin_Xie_Lin,zhao2023masked} increasingly shows that detectors based on the DEtection TRansformer (DETR)\cite{Carion_Massa_Synnaeve_Usunier_Kirillov_Zagoruyko_2020} exhibit superior performance in handling domain gaps compared to those based on Convolutional Neural Networks (CNNs). Unlike purely convolutional designs, DETR innovatively integrates a CNN backbone with transformer architecture (i.e., encoder and decoder), leveraging the attention mechanism to refine features for more detailed representations\cite{Dosovitskiy_Beyer_Kolesnikov_Weissenborn_Zhai_Unterthiner_Dehghani_Minderer_Heigold_Gelly}. Additionally, the transformer architecture has been proven to be highly effective in capturing global structural information \cite{Guo_Wang_Qi_Shi_2023,10388356,10246362}, potentially offering better generalization capability for unseen scenarios. Therefore, it is highly meaningful to explore the effectiveness of DETR-based detectors in addressing the SDG task.

In this paper, we introduce a strong DETR-based detector named the Style-Adaptive DEtection TRansformer (SA-DETR) for SDG in object detection. To better adapt to unseen scenarios, we propose the Online Domain Style Adapter (ODS-Adapter), which addresses the domain gap through statistical matching rather than data augmentation. During training, the adapter employs a dynamic style memory bank to cache the channel-wise statistics (mean and variance) of backbone features and treats them as style bases, consistent with the classic style-transfer paradigm~\cite{huang2017arbitrary}. As iterations progress, similar bases are integrated through momentum-based updates, whereas redundant ones are discarded, allowing the memory bank to self-organize into multiple style bases that faithfully capture the diversity of the training data. During inference, the method first employs the Wasserstein distance\cite{adler2018banach} to quantify distribution discrepancies between the statistics of the testing image and the stored style bases. Next, we convert the distances into weights, which are injected into an AdaIN-like modulation to achieve fast and precise feature projection. Considering that the target domain may exhibit entirely novel styles, the adapter writes the newly observed statistics back to the memory bank during testing, thereby continuously absorbing information from unseen domains and preventing under-adaptation caused by large style gaps.

To further enhance the generalization capability, we have introduced an object-aware contrastive learning module that forces the detector to extract domain-invariant features of objects through contrastive learning. Differing from the proposed adapter that addresses the SDG challenge at the image level, this module employs carefully designed object-aware gating masks to constrain the scope of feature aggregation in both spatial position and semantic category, thereby effectively achieving cross-domain contrast (i.e., between the source domain and the augmented domain) of instance-level features. Specifically, we insert multiple class queries into the transformer encoder, with each query responsible for capturing features of a specific category. Utilizing object-aware gating masks created from annotations, these queries adaptively aggregate specific positional and categorical features from the feature tokens by leveraging a cross-attention mechanism. Then, we introduce a contrastive learning to minimize the differences among these aggregated queries, which can be considered instance-level prototypes for each category. By attracting prototypes of the same category and repelling those of different categories across domains, the detector is forced to extract domain-invariant object features, thereby enhancing its generalization capability.

The main contributions of this paper are as follows:

\begin{itemize}
\item{We show that DETR-based detectors exhibit better generalization in unseen scenarios, and propose a style-adaptive detection transformer. To the best of our knowledge, this is the first work that explores the benefit of a DETR-based detector for the single-source domain generalization task.}

\item{We introduce an Online Domain Style Adapter (ODS-Adapter) that projects the style representation of unseen target domains onto the training domain. By maintaining a dynamic memory bank that self-organizes into diverse style bases and continuously absorbs statistics from unseen domains under a test time adaptation framework, the adapter enables rapid and accurate style adaptation to previously unseen scenarios.}

\item{To further enhance the generalization capability, an Object-aware Contrastive Learning (OCL) module is introduced. This module employs carefully designed object-aware gating masks to constrain the scope of feature aggregation in both spatial position and semantic category, thus forcing the detector to extract domain-invariant features of objects.}
\end{itemize}

Extensive results demonstrate the superior performance and generalization capability of SA-DETR compared to state-of-the-art methods across five different weather conditions. Our results consistently achieved the best outcomes in all scenarios and showed a remarkable \begin{math} 8.4\% \end{math} improvement in mAP on the "Dusk-Rainy" scene.

\section{Related Work}

\subsection{Domain Adaptation and Generalization}

The domain gap is a common issue encountered during network deployment, caused by differences in the distribution between collected data (source domain) and real-world application data (target domain). Current deep learning methods, limited by data-driven supervised learning architectures, often experience significant performance degradation when facing such domain gaps. To address this, domain adaptation techniques\cite{Guan_Huang_Xiao_Lu_Cao_2022,Yu_Aizawa_Irie,Ning_Lu_Xie_Chen_Wei_Zheng_Tian_Yan_Yuan_2023} have been proposed that leverage unlabeled (unsupervised) or minimal (few-shot learning) target domain samples to mitigate the domain gap. Unlike domain adaptation, which assumes access to target domain images during training, the domain generalization task aims to enhance a model's generalization capability to adapt to completely unseen scenarios, making it a more challenging but more practically relevant problem for real-world deployment. In this paper, we address the issue of domain generalization in object detection and focus on using data from a single-source domain.

\subsection{Single-source Domain Generalization for Object Detection}

Single-source Domain Generalization (SDG), a specific case of domain generalization task, aims to enhance network performance on unseen scenarios using only data from one source domain. At present, most research on SDG has concentrated on image classification\cite{Fan_Wang_Ke_Yang_Gong_Zhou_2021,Zhang_Li_Li_Jia_Zhang,10539285,10335732} and segmentation\cite{Huang_Chen_Li_Li_Li_Song_Yan_Xiong,Lee_Seong_Lee_Kim,Li_Li_Li_Guo_2022}, with limited work in the field of object detection, which warrants further exploration. Existing approaches primarily enhance the detector’s generalization capability by implementing data augmentation strategies and feature alignment techniques.

Data augmentation strategies improve model robustness by perturbing images or features to simulate variations that may occur in unseen scenarios. MAD~\cite{Xu_Qin_Chen_Pu_Zhang_2023} transfers input images into the frequency domain and randomizes the distribution of extremely high- and low-frequency components to increase the diversity of source domain data. Building on this, UFR~\cite{liu2024unbiased} incorporates SAM~\cite{Kirillov_Mintun_Ravi_Mao_Rolland_Gustafson_Xiao_Whitehead_Berg_Lo_et} to isolate foreground objects from the background, allowing for targeted augmentation of each component. Liu et al.~\cite{liu2024source} further applies perturbations at both the image and feature levels to simulate a range of color and style biases. CLIP the Gap~\cite{vidit2023clip} leverages the pre-trained vision-language model CLIP~\cite{radford2021learning}, using carefully crafted textual prompts to simulate unknown scene styles and thereby enhance the semantic style features extracted by the backbone. PGST~\cite{li2024phrase} improves the alignment between textual and visual features by leveraging GLIP~\cite{li2022grounded} to project the visual style of source image regions into a hypothetical target domain style. In contrast to the above methods, our proposed adapter projects the style representation of the unseen domain onto the training domain, enabling dynamic style adaptation across diverse unseen scenarios rather than being constrained by the limited sample space of augmented training data.

Feature alignment techniques are usually applied in SDG for object detection to further alleviate the influence of the domain gap. These techniques align cross-domain features by utilizing data from both the source and augmented domains, thereby facilitating the extraction of domain-invariant features in the detector. SDGOD~\cite{wu2022single} introduces a cyclic-disentangled self-distillation framework that achieves domain-invariant feature extraction without annotated labels. MAD~\cite{Xu_Qin_Chen_Pu_Zhang_2023} proposes a Multi-view Adversarial Discriminator to achieve more precise alignment of cross-domain features. To complement the image-level feature alignment methods discussed above, UFR~\cite{liu2024unbiased} has developed a causal prototype learning module that refines features extracted from regions of interest (ROIs) to further mitigate the influence of confounding factors in object attributes. Since the proposed adapter effectively addresses the image-level domain gap, we introduce an object-aware contrastive learning module to further encourage the DETR-based detector to extract domain-invariant instance-level features. This module creates object-aware gating masks to constrain contrastive learning on instance-level features, thereby achieving accurate cross-domain feature alignment for objects.

\subsection{Style Modulation for Domain Adaptation}

Unlike the aforementioned data augmentation approaches, modulating the statistical properties of features provides an effective alternative for style adaptation and has been widely adopted in domain adaptation and generalization tasks. 
IBN-Net~\cite{Pan_Luo_Shi_Tang_2018} and ISW~\cite{Choi_Jung_Yun_Kim_Kim_Choo_2021} first introduced instance normalization and whitening operations to mitigate the influence of domain-specific styles on feature extraction. 
AdaIN~\cite{huang2017arbitrary} was the first to propose the use of channel-wise statistics to modulate feature styles, enabling flexible transfer between arbitrary styles. 
MixStyle~\cite{zhou2021domain} extended this idea to domain generalization in image classification by mixing style statistics across samples to enhance model robustness.
For semantic segmentation, SPC-Net~\cite{Huang_Chen_Li_Li_Li_Song_Yan_Xiong} models style statistics as prototypes, which guide the model’s generalization to unseen domains based on known style representations.
FADA~\cite{bi2024learning} further leverages wavelet transforms to project features into the frequency domain, enabling the decoupling of style and content for more precise modulation. DR-Adapter~\cite{su2024domain} builds on this idea to align diverse target domain styles with the source domain distribution, thereby boosting performance under few-shot conditions.

Recently, style modulation in object detection remains relatively underexplored, often relying on coarse-grained style bases and simplistic modulation processes.
Lopez Rodriguez et al.~\cite{rodriguez2019domain} was the first to apply instance normalization for style unification between source and target domain images, improving feature stability and robustness. 
TDD~\cite{he2022cross} utilizes Fourier-based style mapping to translate source-domain images into target-domain styles and employs supervised learning to enhance the detector’s ability to capture target-specific characteristics.
However, these methods typically require access to target-domain images, making them less applicable to domain generalization scenarios.
To address this issue, FSDA-DETR~\cite{11016953} designed a style projection module to rectify target-domain styles toward the source domain, thereby reducing style shift in few-shot cross-domain detection. LOSA~\cite{11023626} further improved cross-domain adaptability by performing online style projection at both the image and instance levels.
In contrast to the above methods that rely on simple storage and updating of style prototypes, our proposed ODS-Adapter constructs a dynamic style memory bank that self-organizes into a set of discriminative style bases, effectively capturing the diversity of training styles, thereby enabling more precise style modulation.

\subsection{Test-Time Adaptation for Object Detection}

Test-Time Adaptation (TTA) is a learning paradigm that aims to update a pre-trained model during inference using unlabeled target-domain data. 
Tent~\cite{wang2020tent} optimizes batch normalization parameters online by minimizing the entropy loss, thereby enabling the model to adapt to the target domain. 
DUA~\cite{mirza2022norm} eliminates the need for backpropagation and entropy loss, instead achieving adaptation by updating BN statistics in real time. TT-WA~\cite{10833729} dynamically adjusts the affine parameters of the Batch Normalization layers during inference, enabling efficient adaptation to unseen weather conditions.
Further, ActMAD~\cite{mirza2023actmad} models the distribution of entire feature channels to perform more precise feature rectification.

An alternative line of work adopts adaptation through online generated pseudo-labels.
MemCLR~\cite{vs2023towards} employs a mean teacher framework that generates pseudo labels to continuously train the detector on the target domain during inference. AMROD~\cite{cao2024exploring} introduces a dynamic thresholding mechanism to improve pseudo-label quality. MLFA~\cite{10713112} enhances TTA performance by integrating multi-level feature alignment modules into the paradigm.
Differently, we introduce a TTA paradigm designed to address the under-adaptation problem caused by significant discrepancies between training and unseen environments. By continuously absorbing information from the target domain and dynamically updating a memory bank of style representations, our method enables effective adaptation to diverse target domain styles during inference. Since our approach does not rely on generating or using pseudo-labels, it maintains a streamlined, computationally efficient architecture and achieves high adaptation speed.

\section{Method}

\begin{figure*}[htbp]
\centering
\includegraphics[width=18cm]{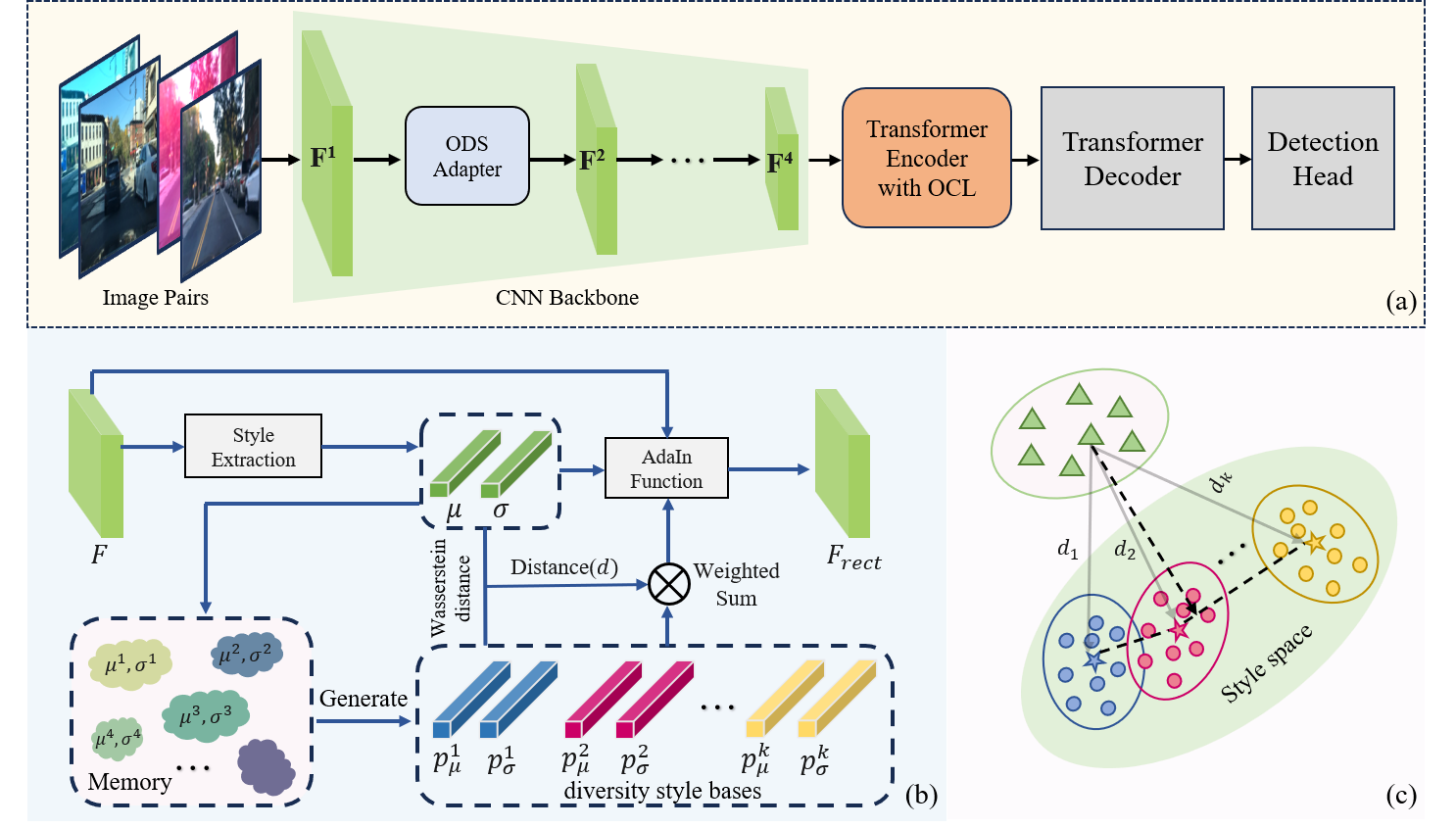}
\caption{
Details of (a) the overview of the proposed SA-DETR framework, which adopts DINO as the base detector and incorporates an Online Domain Style Adapter (ODS-Adapter) and an Object-aware Contrastive Learning (OCL) module, and (b) the workflow of the ODS-Adapter. Meanwhile, the memory module follows a test‑time adaptation architecture: it operates during both training and inference to store the statistics of input features and construct multiple style bases, as indicated by the blue arrows. (c) illustrating the projection process during inference. The adapter rectifies the channel wise statistics of the feature maps extracted by the backbone using the stored style bases, thereby mapping the style of the unseen target domain onto that of the training domain.
}
\label{fig_1}
\end{figure*}

SA-DETR is a strong DETR-based detector for single-source domain generalization in object detection. In this section, we first provide a framework overview, which includes the DETR revisit, the data augmentation method and the framework of our detector.  Then, we detail the proposed domain style adapter and the object-aware contrastive learning module in subsequent sections. Finally, the loss functions used in our method are described in the overall training objective section.

\subsection{Framework Overview}

\subsubsection{DETR revisit} In our approach, we utilize DINO\cite{zhang2022dino} as the base detector, an advanced DETR-based end-to-end object detector. Specifically, DINO incorporates a CNN backbone for feature extraction and employs an encoder-decoder transformer structure along with a detection head for the final detection predictions. When processing an image, the backbone initially extracts multi-scale feature maps similar to those used in CNN-based detectors. These feature maps are then reshaped into feature tokens that feed into the encoder-decoder module. The encoder further refines the input features by using self-attention. The decoder uses inserted object queries to probe local regions and aggregate instance features from the refined features through cross-attention, generating output embeddings. Finally, these output embeddings are used to compute prediction results through the detection head.

\subsubsection{Data augmentation} The objective of single-source domain generalization is to train the network with data from a single domain and to generalize its performance to a variety of unseen scenarios. In this task, data augmentation methods are essential to break the limitations of sample diversity. Inspired by previous work\cite{Xu_Qin_Chen_Pu_Zhang_2023}, we employ spurious correlation generator to augment the source dataset at the image level, ultimately yielding an augmented dataset of equivalent size. Therefore, the input to our method includes both the original and augmented data. Building upon this foundation, we propose the ODS-adapter to leverage known styles for representing unseen ones, thereby overcoming the limitations of augmentation methods in improving performance.

\subsubsection{Framework} Fig. \ref{fig_1}, Our SA-DETR adopts DINO as the base detector and integrates our designed methods, including an ODS-Adapter and an OCL module. Specifically, during training, the network inputs consist of pairs of images, each pair containing an equal number of images from both the source and augmented domains. For a batch of image pairs, the detector first uses a ResNet backbone\cite{He_Zhang_Ren_Sun_2016} to extract multi-scale feature maps, and feeds the maps from each layer into our proposed adapter. The adapter leverages a dynamic style memory bank to cache the channel-wise statistics of backbone features and to construct multiple style bases. Then, we insert an OCL module into the encoder during training, which utilizes contrastive learning to encourage the extraction of instance-level domain-invariant features, thereby enhancing the detector's generalization capabilities. Ultimately, the refined features are input into the subsequent decoder and detection head to obtain detection results, in line with DINO. During inference, the adapter rectifies the channel‑wise statistics of the feature maps extracted by the backbone using the stored style bases, thereby projecting the style of the unseen target domain onto that of the training domain. Notably, to mitigate under‑adaptation arising from large discrepancies between the training and unseen environments, our adapter follows a TTA framework that continuously absorbs target domain information and dynamically updates a memory bank of style bases. Our OCL module is used only during training as a plug in to facilitate the detector’s representation learning. More details on the ODS-Adapter and the OCL module will be presented in Subsections \ref{Subsection B} and \ref{Subsection C}, respectively.

\subsection{Online Domain Style Adapter}
\label{Subsection B}

We present ODS-Adapter that dynamically projects target domain style representations onto the training domain manifold, enabling robust adaptation to diverse unseen scenarios. The adapter first constructs a dynamic style memory bank that self-organizes into multiple discriminative style bases. At inference, it applies a weighted AdaIN projection to achieve fine grained style modulation, thereby enhancing detector robustness on unseen domains. Moreover, ODS-Adapter supports test time adaptation by continually updating the memory bank, effectively mitigating mismatches between training and deployment environments.

Given the feature map of a specific layer \begin{math} F\in\mathbb{R}^{(B\times C\times H\times W)} \end{math}, we first compute its statistical data, i.e., the mean \begin{math}\mu \end{math} and variance \begin{math} \sigma \end{math} along the channel dimension, as follows:
\begin{equation}
\mu\left(F\right) =\frac{1}{H W} \sum_{h=1}^{H} \sum_{w=1}^{W} F,
\end{equation}
\begin{equation}
\sigma\left(F\right) =\sqrt{\frac{1}{H W} \sum_{h=1}^{H} \sum_{w=1}^{W}\left(F-\mu\left(F\right)\right)^{2}+\epsilon},
\end{equation}
where \begin{math}\mu \end{math} and \begin{math} \sigma \in\mathbb{R}^{(B\times C)}\end{math}, \begin{math}\epsilon \end{math} is set to \begin{math}1 \times 10^{-6}\end{math} to avoid numerical computation resulting in zero. In this manner, we obtain channel-wise statistics for the feature maps of each layer, which are subsequently leveraged to construct the style memory bank and to drive the ensuing style adaptation procedure.

Next, we design a dynamic style memory bank that stores the statistics extracted at every training iteration and supports on-the-fly style adaptation during inference. The bank is implemented as a ring buffer that maintains \begin{math} K \end{math} style prototypes and is updated after each mini batch. Concretely, the mean and deviation of every incoming feature map are compared with all stored prototypes, and their style discrepancy is measured by the Wasserstein distance, as follows:
\begin{equation}
d_i\bigl(u,\, p_u^{\,i},\, \sigma,\, p_\sigma^{\,i}\bigr)
  = \left\lVert u - p_u^{\,i} \right\rVert_{2}^{2}
    + \bigl(\sigma^{2} + (p_{\sigma}^{\,i})^{2} - 2\sigma\, p_\sigma^{\,i}\bigr),
\end{equation}
where \begin{math} d_i \end{math} measures the stylistic similarity between the input feature map and the \begin{math}i\end{math}-th stored prototype. The terms \begin{math} p_u^{\,i} \end{math} and \begin{math} p_\sigma^{\,i} \end{math} denote the mean and variance of the \begin{math}i\end{math}-th prototype in the memory bank, respectively.

To dynamically fuse similar style prototypes while retaining the most discriminative ones, thereby enabling the memory bank to self-organize a set of discriminative style bases, we introduce an adaptive threshold. \begin{math}\tau\end{math}. The threshold is updated online based on the distance distribution of the current mini batch, and it is computed as follows:
\begin{equation}
\tau \;=\; \frac{\alpha}{K}\,\sum_{i=1}^{K} d_i
\end{equation}
where \begin{math}\alpha\end{math} is a temperature coefficient that controls the sensitivity of the adaptive threshold. Let \begin{math}d_{\min}= \min_{k}(d_{k})\end{math} denote the distance between the current sample and its closest prototype. If \begin{math}d_{\min}> \tau\end{math}, the sample exhibits a significant style discrepancy. In that case, we follow the queue update strategy~\cite{9157636} to locate the least frequently used prototype in the memory bank and replace it with the statistics of the current sample. Otherwise, we fuse the current statistics with the nearest prototype by applying an Exponential Moving Average (EMA)~\cite{Marsella_Gratch_2009}, thereby updating its mean and standard deviation to preserve the continuity and stability of the bank, as detailed below:
\begin{equation}
p_{\mu}'=\lambda p_{\mu}+(1-\lambda) \mu,
\end{equation}
\begin{equation}
p_{\sigma}'=\lambda p_{\sigma}+(1-\lambda) \sigma,
\end{equation}
where \begin{math}\lambda\end{math} denotes the momentum factor. Our memory bank follows a test-time adaptation design. During testing, we employ the same update strategy to enable the model to rapidly and continuously absorb the style characteristics of unseen scenes. It is noteworthy that only the fusion is employed to mitigate the contamination of the memory bank caused by discrete or anomalous samples during the TTA phase.

Finally, we employ the style prototypes stored in the memory bank to rectify the input image by mapping the target domain statistics to the training domain distribution. To maintain consistency with previous stages, we compute the weighting coefficients using the Wasserstein distance between the current statistics and each style prototype. Additionally, we explore a soft-KNN retrieval mechanism that selects a small set of representative bases instead of utilizing the entire bank, which is found to be suboptimal. Therefore, the resulting distances \begin{math} d=\{d_1,d_2,...,d_K\} \end{math} are normalized using softmax, so that the weights sum to 1. These normalized weights act as coefficients in AdaIN, steering the subsequent feature projection as follows:
\begin{equation}
{\mu}'=softmax(d )\cdot p_{\mu},{\sigma}'=softmax(d )\cdot p_{\sigma},
\end{equation}
\begin{equation}
F_{rect} = \frac{F-\mu }{\sigma } {\sigma}' +{\mu }' ,
\end{equation}
where \begin{math} F \end{math} represents the original feature map of the input image, \begin{math} F_{rect} \end{math} corresponds to its rectified feature map, "\begin{math} \cdot \end{math}" denotes matrix multiplication. We rectify every layer of the extracted multiscale feature maps during both the training and inference phases to ensure a consistent workflow in the detector.

\subsection{Object-aware Contrastive Learning Module}
\label{Subsection C}

\begin{figure*}[htbp]
\centering
\includegraphics[width=18cm]{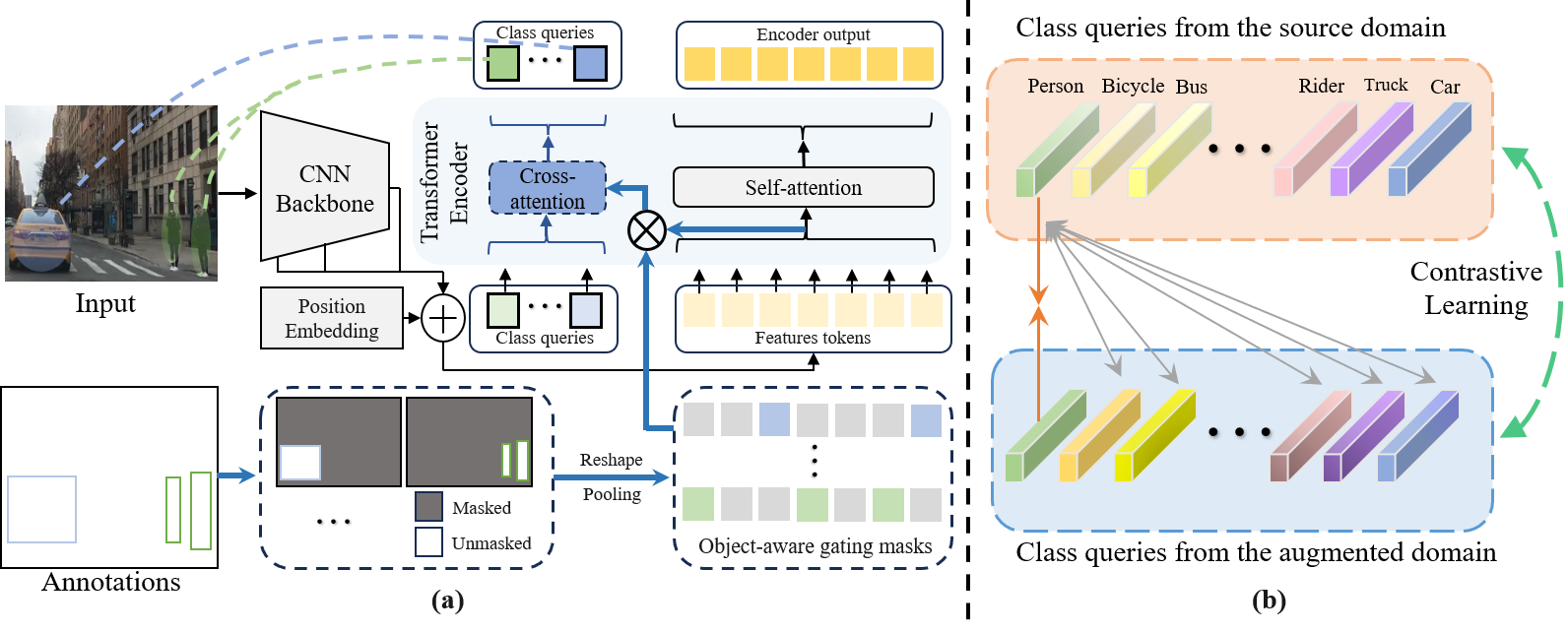}
\caption{
Our proposed object-aware contrastive learning (OCL) module. Details of (a) the inserted class queries adaptively capture features of objects by utilizing the generated object-aware gating masks, and (b) cross-domain instance-level feature alignment is achieved by leveraging contrastive learning.}
\label{fig_2}
\end{figure*}

Since DETR-based detectors lack region proposals, we leveraged the characteristics of the single-source domain generalization task, where both the source and augmented domains are labeled, to generate object-aware gating masks. By using these masks to filter out the aggregation of irrelevant features, we can accurately implement cross-domain contrast (i.e., between the source domain and the augmented domain) of instance-level features. The module structure is shown in Fig. \ref{fig_2}.

\subsubsection{Object-aware gating masks} We used the bounding boxes and their corresponding classes from the annotations to generate object-aware gating masks. Specifically, let \begin{math} Y = \{(b_1,c^1),(b_2,c^2),...,(b_m,c^m)\}\end{math} represent the annotations, where \begin{math} b_m \end{math} and \begin{math} c^m \end{math} denote the bounding box and category of the \begin{math} m \end{math}-th object, respectively. We first aggregate all instances of the same category to obtain the coordinate set \begin{math} B^c = \{b^c_1,b^c_2,...,b^c_i\}\end{math}, where each \begin{math} b \end{math} includes two coordinates represented as \begin{math} (x_{min},y_{min},x_{max},y_{max}) \end{math}. Then, we generate the gating masks based on the coordinate set \begin{math} B^c \end{math} as follows:
\begin{equation}
g_{x,y}^{c}=\left\{\begin{array}{ll}
1, & \text { if } x_{min}\le x \le x_{max} \ \text {and } y_{min}\le y \le y_{max};\\
0, & \text { otherwise. }
\end{array}\right.
\end{equation}

We use \begin{math} G^c=\{g_{x,y}^c\}_{H\times W } \end{math} to denote the gating mask corresponding to category \begin{math} c \end{math}, where \begin{math} g_{x,y}^c=1 \end{math} indicates that the current pixel position contains the object, and occlusion calculation of the attention mechanism is not required. Finally, we need to perform multi-scale downsampling and concatenation on the obtained gating masks to ensure alignment with the positions of the feature tokens input into the encoder.

\subsubsection{Instance-level feature aggregation with class queries} We insert \begin{math} c \end{math} class queries \begin{math} Q=\{{q}_1,q_2,\ldots,q_c\} \end{math} into the encoder, with each query \begin{math} q_i\in\mathbb{R}^d \end{math} being a learnable vector designed to aggregate features of a specific category using a cross-attention mechanism. The scope of aggregation is controlled by the object-aware gating masks previously obtained, enabling aggregation solely for instance-level features. Specifically, we define the feature tokens as \begin{math} T \end{math}, with corresponding position encodings \begin{math} P \end{math}, and object-aware gating masks as \begin{math} M=\{{G}^1,G^2,\ldots,G^c\} \end{math}. In each transformer block of the encoder, all class queries \begin{math} Q \end{math} are updated in parallel through cross-attention with the feature tokens, as follows:
\begin{equation}
Q^l=Q^{l-1}+CrossAtten(Q^{l-1},T^{l-1}+P,T^{l-1},M),
\end{equation}
where CrossAtten(query, key, value, key\_padding\_mask) denotes the standard cross-attention layer, the superscript \begin{math} l \end{math} represents the \begin{math} l \end{math}-th block of the encoder. Ultimately, we obtain the class queries \begin{math} Q \end{math} that aggregate instance-level features of each category.

\subsubsection{Contrastive learning for feature alignment} Since the input images to our method include both the source and augmented domains, we can obtain class queries from two distinct domains. By attracting positive sample pairs (queries from the same category across both domains) and repelling negative samples, the detector is compelled to extract domain-invariant features from objects. Specifically, we define \begin{math} Q^S \end{math} and \begin{math} Q^A\in\mathbb{R}^{C\times d} \end{math} as class query sets from the source and augmented domains, respectively. The contrastive learning loss is constructed as follows:
\begin{equation}
\begin{aligned}
L_{contra}\ =-\frac{1}{C}\sum_{i=1}^{C}{log\ \frac{exp\left(q_i^S \cdot q_i^A\right)}{\sum_{j=1}^{C}exp\left(q_j^S \cdot q_i^A\right)}},
\end{aligned}
\end{equation}
where "\begin{math} \cdot \end{math}" is used to measure the similarity between queries from different domains, and \begin{math} C \end{math} represents the total number of categories in the dataset. Meanwhile, we compute the loss only for the categories that are present in the images.

\subsection{Overall Training Objective}

Since our OCL module employs contrastive learning to further enhance the generalization capability of the detector, the SA-DETR is trained using two types of loss functions: the supervised detection loss \begin{math} L_{det} \end{math}, which is consistent with the DINO, and the contrastive learning loss \begin{math} L_{contra} \end{math} as defined in Eq. (11). The training objective can be defined as follows:
\begin{equation}
L = L_{det} +  \lambda_{c}L_{contra},
\end{equation}
where \begin{math} L_{det} \end{math} and \begin{math} L_{contra} \end{math} are used for both the source images and the augmented images, and \begin{math} \lambda_{c} \end{math} denotes the balancing weights for the corresponding learning loss functions.

\section{EXPERIMENTS}

This section details our experiments, including Datasets, Implementation Details and Evaluation Metrics, Ablation Studies, Comparisons with State-of-the-Art, and Visualization and Analysis. Detailed discussions of each aspect are provided in the following subsections.


\renewcommand{\arraystretch}{1.25}  
\begin{table*}[t]
  \centering
  \small
  \caption{Ablation study of SA-DETR. All numbers are $mAP_{50}$ (\%).  
          }
  \label{tab:Ablation}

  \begin{tabular}{l c c c c |c|c c c c}
    \toprule
    \multirow{2}{*}{Model Component} & \multirow{2}{*}{Aug.} &
    \multicolumn{2}{c}{ODS‑Adapter} & \multirow{2}{*}{OCL} &
    \textbf{Source domain} &
    \multicolumn{4}{c}{\textbf{Unseen target domains}} \\
    \cmidrule(lr){3-4}\cmidrule(lr){6-10}   
    & & Std. & TTA & & Day‑C & Day‑F & Dusk‑R & Night‑R & Night‑C \\  
    \midrule
    vanilla DINO        &            &            &            &            & 61.6 & 38.0 & 35.7 & 17.3 & 43.7 \\
    +Aug                &\checkmark  &            &            &            & 63.8 & 41.3 & 37.5 & 18.6 & 44.5 \\
    +Adapter            & \checkmark  & \checkmark  &            &            & 64.2 & 41.7 & 45.1 & 20.7 & 45.0 \\
    +Adapter \& TTA      & \checkmark  & \checkmark  & \checkmark  &            & 64.3 & \textbf{42.1} & 45.3 & 22.9 & 45.4 \\
        SA‑DETR      & \checkmark  & \checkmark  & \checkmark  & \checkmark  &\textbf{64.3} & 41.7 & \textbf{46.5} & \textbf{24.5} & \textbf{45.6} \\
    \bottomrule
  \end{tabular}
\end{table*}

\subsection{Datasets}

Following\cite{wu2022single,liu2024unbiased}, we employed a benchmark for single-source domain generalization in object detection to evaluate the SA-DETR. The dataset consists of scenes under five different weather conditions, meticulously selected from three public datasets: Berkeley Deep Drive 100K (BDD-100K)\cite{yu2020bdd100k}, Cityscapes\cite{Cordts_Omran_Ramos_Rehfeld_Enzweiler_Benenson_Franke_Roth_Schiele_2016}, and Adverse-Weather\cite{Hassaballah_Kenk_Muhammad_Minaee_2021}. Specifically, the experiment designated "Daytime-Clear" as the source domain, which includes 19,395 training images and 8,313 testing images. The other four conditions, used exclusively for testing as unseen scenarios, included "Daytime-Foggy" with 3,775 test images, "Dusk-Rainy" with 3,501 test images, "Night-Clear" with 26,158 test images, and "Night-Rainy" with 2,494 test images. All datasets encompassed seven categories: person, car, bike, rider, motor, bus, and truck.

\subsection{Implementation Details and Evaluation Metric}
Our approach is based on the DINO detector and is compared with existing methods. To ensure a fair comparison, we use ResNet-50 \cite{He_Zhang_Ren_Sun_2016} as the backbone. In all experiments, our method employed a 1x training configuration, was trained in just 12 epochs, and achieved outstanding performance. For all scenarios, we set the weight factors \begin{math} \lambda_{c} \end{math} in Eq. (12) to 0.1. Consistent with DINO's settings, we trained our models using the Adam optimizer\cite{Kingma_Ba_2014} with a base learning rate of \begin{math} 2\times 10^{-4} \end{math}, which was reduced by a factor of 0.1 in the 11-th epoch. We conduct each experiment on an NVIDIA A6000 GPU with 48 GB of memory.

We use the mean Average Precision (mAP) to evaluate our method, as follows:
\begin{equation}
\mathrm{mAP}=\frac{1}{c} \sum_{i=0}^{c} \mathrm{AP}_{i},
\end{equation}
where \begin{math} \mathrm{AP}=\int_{0}^{1} P(R) d R \end{math} represents the area under the precision-recall curve and \begin{math} c \end{math} denotes the number of categories. Consistent with the evaluation of other methods, we calculate the object detection evaluation metric at an IoU threshold of \begin{math} 50\% \end{math}.

\subsection{Ablation Studies}
\label{Subsection Ablation}

In this section, we first conduct ablation studies by selectively removing individual components to quantify their contributions to our framework. Next, we focus on the proposed ODS-Adapter, analyzing the effect of the number of stored style bases and the role of the constructed dynamic memory bank. Finally, we report the runtime cost of style rectification to highlight the efficiency of our approach.

\subsubsection{Effectiveness of Each Component} Table~\ref{tab:Ablation} reports the ablation study, highlighting the contribution of each component. As shown in rows 1 and 2 of the table, "Aug." refers to the data-augmentation method~\cite{Xu_Qin_Chen_Pu_Zhang_2023}, which yields only marginal gains in challenging conditions such as "Dusk-Rainy" and "Night-Rainy". These results confirm that conventional augmentation alone is insufficient when test scenes deviate substantially from the training domain. The results in row 3 show that adding our ODS-adapter markedly improves performance on these hard domains while maintaining the accuracy of the source domain. Enabling the adapter in test-time adaptation mode further raises "Night-Rainy" mAP to 22.9\% and increases the average target domain mAP, demonstrating that online prototype updates effectively complement the offline style learned during training.

To further enhance the generalization of the detector, we propose the OCL module. This module improves the detector's generalization by encouraging the extraction of domain-invariant instance features. As shown in row 5, the detector's performance is further improved across various scenarios by incorporating the OCL module. However, we observe a modest performance drop in the "Daytime-Foggy" test scenes that closely resemble the training domain, possibly because subtle domain specific cues such as slight contrast variations assist the detector in localizing objects. Ultimately, our SA-DETR achieves significant improvements in all experimental scenarios compared to the vanilla DINO detector, demonstrating that the proposed components effectively enhance the detector's generalization.

\subsubsection{Impact of the Number of Stored Style Bases} Table \ref{tab:num_prototypes} investigates the effect of the number of stored style bases on detection performance. When \begin{math} k=1 \end{math}, all style statistics are collapsed into a single prototype, similar to existing methods \cite{su2024domain,11016953,11023626}, resulting in a sub-optimal average $mAP_{50}$ of 42.72\%. Increasing \begin{math} k \end{math} from 1 to 4 steadily improves performance, with the largest gain appearing on the most challenging domains and a peak average of 44.52\% at \begin{math} k=4 \end{math}. Beyond this point, adding more bases provides no additional benefit and even causes a slight decline, likely because single source training data do not offer enough stylistic diversity to populate a larger memory without redundancy or noise. These results demonstrate that the robustness of our adapter enhances the detector’s ability to adapt to unseen scenes. A modest number of bases further improves performance by striking an optimal balance between capturing meaningful intra-domain style variations and avoiding over fragmentation of the memory.

\begin{table}[t]
  \centering
  \small
  \setlength{\tabcolsep}{5pt}
  \caption{Impact of the number of stored style bases on $mAP_{50}$ (\%).}
  \label{tab:num_prototypes}

  \begin{tabular}{c|ccccc|c}
    \toprule
    Number & Day‑C & Day‑F & Dusk‑R &  Night‑R &Night‑C & Avg \\
    \midrule
    1 & 64.0 & 41.8 & 42.8 & 19.8 & 45.2 & 42.72 \\
    2 & 64.0 & 42.2 & 44.1 & 23.8 & 44.9 & 43.80 \\
    3 & 63.7 & 41.7 & 45.0 & 24.0 & 45.2 & 43.92 \\
    4 & 64.3 & 41.7 & \textbf{46.5} & \textbf{24.5} & \textbf{45.6} & \textbf{44.52} \\
    5 & \textbf{64.5} & \textbf{42.3} & 45.7 & 23.2 & 45.5 & 44.24 \\
    6 & 64.2 & 42.2 & 45.4 & 22.2 & 45.6 & 43.92 \\
    8 & 64.5 & 41.5 & 45.5 & 22.5 & 45.4 & 43.88 \\
    \bottomrule
  \end{tabular}
\end{table}

\subsubsection{Hyperparameter sensitivity of the dynamic memory bank} As shown in Table \ref{tab:memory_bank}, we further validate the effectiveness of the self-organizing strategy, which discards rarely used style bases and allows the memory bank to retain a more discriminative set of style bases. “N/A” denotes using only EMA to fuse the computed statistics, which is clearly suboptimal. On the other hand, we analyze the sensitivity of the dynamic memory bank to $\alpha$ and observe only minor performance fluctuations. The best average $mAP_{50}$ is achieved when $\alpha = 0.7$, largely due to more substantial gains on challenging low-light domains such as Dusk-R and Night-R.

\begin{table}[t]
  \centering
  \small
  \setlength{\tabcolsep}{4pt}
  \renewcommand{\arraystretch}{1.05}
  \caption{Effect of the temperature coefficient $\alpha$ in the dynamic memory bank on $mAP_{50}$ (\%).}
  \label{tab:memory_bank}
  \begin{tabular}{lccccc|c}
    \toprule
    $\alpha$ & Day-C & Day-F & Dusk-R & Night-R &Night-C  & Avg. \\
    \midrule
    N/A    & 64.4 & 42.0 & 45.2 & 23.4 & 45.3 & 44.06 \\
    \midrule
    0.6    & \textbf{64.7} & 42.3 & 45.1 & 24.3 & \textbf{45.9} & 44.46 \\
    0.7    & 64.3 & 41.7 & \textbf{46.5} & \textbf{24.5} & 45.6 & \textbf{44.52} \\
    0.8    & 64.4 & \textbf{42.5} & 45.4 & 23.9 & 45.3 & 44.30 \\
    \bottomrule
  \end{tabular}
\end{table}

\subsubsection{Analysis of inference efficiency} Our proposed ODS-Adapter enables style rectification for unseen domains during inference and also performs dynamic updates to the style bases stored in a memory bank within a TTA framework. To evaluate the additional runtime overhead introduced during inference, we measure the average inference time over 500 runs to ensure the stability of the results, as reported in Table \ref{tab:inf_time}. After incorporating the Adapter, the average inference time is 87.4 ms, which introduces only a slight runtime overhead to the vanilla DINO. This modest increase in inference time can be seen as a reasonable trade-off for the performance gains brought by style rectification and dynamic style memory updates. Since we only dynamically update the style representation memory bank during testing, it not only further enhances generalization to unseen scenes but also incurs virtually no additional time overhead, demonstrating the effectiveness and efficiency of our ODS-Adapter. It is worth noting that the proposed OCL module is only used to assist the detector during training and can be removed at inference time, thus incurring no additional overhead.

\begin{table}[hb]
  \centering
  \small
  \caption{Average inference time per image (ms).}
  \label{tab:inf_time}

  \begin{tabular}{l c}
    \toprule
    Method & {Time / ms} \\
    \midrule
    DINO                  & 81.2 \\
    DINO + Adapter        & 87.4 \\
    DINO + Adapter \& TTA  & 88.3 \\
    \bottomrule
  \end{tabular}
\end{table}

\subsection{Comparisons with the State-of-the-Art}

Following \cite{wu2022single,liu2024unbiased}, all methods are trained on the “Daytime-Clear” dataset and tested on five different weather scenarios. In each scenario, we compare our method with state-of-the-art SDG methods to demonstrate its effectiveness and generalization capability. Furthermore, we show the performance of the vanilla DINO to highlight the superior generalization of DETR-based detectors in some unseen scenarios.

\begin{table}[ht]
  \centering
  \caption{Experimental results (\%) on Daytime-Clear Scene.}
   \scalebox{1.0}{%
\begin{tabular}{c@{\hskip 4pt}|c@{\hskip 4pt}c@{\hskip 4pt}c@{\hskip 4pt}c@{\hskip 4pt}c@{\hskip 4pt}c@{\hskip 4pt}c@{\hskip 4pt}|>{\columncolor{gray!20}}c}
    \toprule
    Methods & Bus   & Bike  & Car   & Motor & Person & Rider & Truck & $mAP_{50}$ \\
    \midrule
    Source-FR\cite{Ren_He_Girshick_Sun_2017} & 66.9      &  45.9     & 69.8      &46.5       & 50.6      & 49.4      & 64.0      &56.2  \\
    SW\cite{Pan_Zhan_Shi_Tang_Luo_2019}    & 62.3      & 42.9      &53.3       &49.9       & 39.2      &46.2       &60.6       &50.6  \\
    IBN-Net\cite{Pan_Luo_Shi_Tang_2018} &63.6       &40.7       &  53.2     &45.9       &38.6       &45.3       &60.7       &49.7  \\
    IterNorm\cite{Huang_Zhou_Zhu_Liu_Shao_2019} & 58.4      & 34.2      &42.4       &44.1  &31.6       & 40.8      &55.5       &43.9  \\
    ISW\cite{Choi_Jung_Yun_Kim_Kim_Choo_2021}   &62.9       &44.6       & 53.5      &49.2       & 39.9      &48.3       & 60.9      & 51.3 \\
    SDGOD\cite{wu2022single} & \textbf{68.8}      & 50.9      & 53.9      &\textbf{56.2}      & 41.8      &52.4       & \textbf{68.7}      &56.1  \\
    CLIP-Gap\cite{vidit2023clip} & 55.0      &47.8       &67.5       & 46.7      & 49.4      &46.7       & 54.7      & 52.5 \\
    DivAlign\cite{10656467}   & -      & -     &-       & -      & -     &-      &-     &52.8  \\
    UFR\cite{liu2024unbiased}   & 66.8      & 51.0      & 70.6      &55.8       &49.8       & 48.5      &67.4       &58.6  \\
    \midrule
    \midrule
    DINO\cite{zhang2022dino} &62.6       &50.7       &84.3       &50.8       &66.5       &51.7       &64.5       &61.6  \\
    SA-DETR & 63.8      &\textbf{55.8}      & \textbf{85.1}      &51.9      & \textbf{68.9}     &\textbf{57.2}       & 66.4      &\textbf{64.3}  \\
    \bottomrule
    \end{tabular}%
  }
  \label{tab:Daytime-Clear}%
\end{table}%

\subsubsection{Results on Daytime-Clear Scene} We first evaluate the model's performance on the source domain. As shown in Table \ref{tab:Daytime-Clear}, the vanilla DINO naturally surpasses CNN-based detectors, and our method further achieves a \begin{math} 64.3\% \end{math} mAP, outperforming the state-of-the-art method by \begin{math} 5.7\% \end{math}. This can be attributed to the strong ability of the DETR-based detector, as well as the effective data augmentation and the proposed OCL module, which further enhance the detector's performance.

\subsubsection{Results on Dusk-Rainy and Night-Rainy Scene} We evaluated our method's performance in two unseen scenarios: “Dusk-Rainy” and “Night-Rainy”, with results presented in Tables \ref{tab:Dusk-Rainy} and \ref{tab:Night-Rainy}. In the “Dusk-Rainy” scenario, our method achieves a 46.5\% mAP, surpassing the optimal result by 8.4\%. In the Night-Rainy scenario, the vanilla DINO performs below expectations, indicating that the existing domain gap between training and testing data indeed significantly reduces the detector's accuracy. Meanwhile, our method also demonstrated outstanding performance, achieving a \begin{math} 24.5\% \end{math} mAP, which exceeds the state-of-the-art method by a margin of \begin{math} 0.4\% \end{math} in mAP. This shows that the proposed ODS-Adapter effectively enhances the detector's robustness against significant distribution differences between the training and testing scenarios.

\begin{table}[tbp]
  \centering
  \caption{Experimental results (\%) on Dusk-Rainy Scene.}
   \scalebox{1.0}{%
\begin{tabular}{c@{\hskip 4pt}|c@{\hskip 4pt}c@{\hskip 4pt}c@{\hskip 4pt}c@{\hskip 4pt}c@{\hskip 4pt}c@{\hskip 4pt}c@{\hskip 4pt}|>{\columncolor{gray!20}}c}
    \toprule
    Methods & Bus   & Bike  & Car   & Motor & Person & Rider & Truck & $mAP_{50}$ \\
    \midrule
    FR\cite{Ren_He_Girshick_Sun_2017} &34.2     &21.8    &47.9      &16.0      &22.9     &18.5      &34.9     &28.0  \\
    SW\cite{Pan_Zhan_Shi_Tang_Luo_2019}    &35.2      &16.7       &50.1       &10.4       &20.1      &13.0       &38.8       &26.3   \\
    IBN-Net\cite{Pan_Luo_Shi_Tang_2018} &37.0       &14.8       &50.3       &11.4       & 17.3      &13.3       & 38.4      &26.1   \\
    IterNorm\cite{Huang_Zhou_Zhu_Liu_Shao_2019} & 32.9      & 14.1      &38.9       &11.0       &15.5       &11.6       &35.7       &22.8   \\
    ISW\cite{Choi_Jung_Yun_Kim_Kim_Choo_2021}   &34.7       &16.0       &50.0       &11.1       &17.8       &12.6       &38.8       & 25.9   \\
    SDGOD\cite{wu2022single} &37.1       &19.6       &50.9       &13.4       &19.7       &16.3       &40.7       &28.2   \\
    CLIP-Gap\cite{vidit2023clip} &37.8       &22.8       &60.7       &16.8       &26.8       &18.7       &42.4       &32.3   \\
    DivAlign\cite{10656467}   & -      & -     &-       & -      & -     &-      &-     &38.1  \\
    UFR\cite{liu2024unbiased}   &37.1       &21.8       &67.9       &16.4       &27.4       &17.9       &43.9       &33.2   \\
    SRCD\cite{10742956}   &39.5      &21.4    &50.6       &11.9       &20.1      &17.6       &40.5      &28.8   \\
    \midrule
    \midrule
    DINO\cite{zhang2022dino} &  40.6     &23.8       &73.0       &14.2       &34.6       &18.4       &45.0       &35.7   \\
    SA-DETR &\textbf{48.8}       &\textbf{33.0}       &\textbf{77.6}       &\textbf{30.9}       &\textbf{50.2}       &\textbf{31.4}       &\textbf{53.2}       &\textbf{46.5}   \\
    \bottomrule
    \end{tabular}%
  }
  \label{tab:Dusk-Rainy}%
\end{table}%

\begin{table}[htbp]
  \centering
  \caption{Experimental results (\%) on Night-Rainy Scene.}
   \scalebox{1.0}{%
\begin{tabular}{c@{\hskip 4pt}|c@{\hskip 4pt}c@{\hskip 4pt}c@{\hskip 4pt}c@{\hskip 4pt}c@{\hskip 4pt}c@{\hskip 4pt}c@{\hskip 4pt}|>{\columncolor{gray!20}}c}
    \toprule
    Methods & Bus   & Bike  & Car   & Motor & Person & Rider & Truck & $mAP_{50}$ \\
    \midrule
    Source-FR\cite{Ren_He_Girshick_Sun_2017} &21.3     &7.7     &28.8      &6.1      &8.9       &10.3      &16.0     &14.2  \\
    SW\cite{Pan_Zhan_Shi_Tang_Luo_2019}    &22.3      &7.8       &27.6       &0.2       &10.3      &10.0       & 17.7      &13.7   \\
    IBN-Net\cite{Pan_Luo_Shi_Tang_2018} &24.6       &10.0       &28.4       &0.9       &8.3       &9.8       &18.1       &14.3   \\
    IterNorm\cite{Huang_Zhou_Zhu_Liu_Shao_2019} &21.4       &6.7       &22.0       &0.9       &9.1       &10.6       &17.6       &12.6   \\
    ISW\cite{Choi_Jung_Yun_Kim_Kim_Choo_2021}   &22.5       &11.4       &26.9       &0.4       &9.9       &9.8       &17.5       &14.1   \\
    SDGOD\cite{wu2022single} &24.4       &11.6       &29.5       &\textbf{9.8}       &10.5       &11.4       &19.2       & 16.6  \\
    CLIP-Gap\cite{vidit2023clip} &28.6       &12.1       &36.1       &9.2       &12.3       &9.6       &22.9       &18.7   \\
    DivAlign\cite{10656467}   & -      & -     &-       & -      & -     &-      &-     &24.1  \\
    UFR\cite{liu2024unbiased}   &29.9       &11.8       &36.1       &9.4       &13.1       &10.5       &23.3       &19.2   \\
    SRCD\cite{10742956}   &26.5      &12.9    & 32.4      &0.8       &10.2      &\textbf{12.5}      & 24.0      &17.0   \\
    \midrule
    \midrule
    DINO\cite{zhang2022dino} &28.6       &10.4       &39.9       &0.2       &12.5       &7.4       &22.2       &17.3   \\
    SA-DETR &\textbf{38.9}       &\textbf{12.9}       &\textbf{51.3}       &0.7       &\textbf{26.1}       &11.7       &\textbf{30.0}      &\textbf{24.5}   \\
    \bottomrule
    \end{tabular}%
  }
  \label{tab:Night-Rainy}%
\end{table}%

\subsubsection{Results on Daytime-Foggy and Night-Clear Scene} Tables \ref{tab:Daytime-Foggy}and \ref{tab:Night-Clear} show the results of methods in two other scenarios that have smaller stylistic differences compared to those discussed above. The vanilla DINO also achieved competitive results, demonstrating that DETR-based detectors have better generalization for various unseen scenarios. Our method achieves the best performance, surpassing the state-of-the-art methods by \begin{math} 2.1\% \end{math} and \begin{math} 3.1\% \end{math} in mAP, respectively. These improvements are primarily achieved through the combined effects of our adapter and OCL module. The experiments demonstrate SA-DETR's powerful capability in addressing SDG object detection tasks.

\begin{table}[tp]
  \centering
  \caption{Experimental results (\%) on Daytime-Foggy Scene.}
   \scalebox{1.0}{%
\begin{tabular}{c@{\hskip 4pt}|c@{\hskip 4pt}c@{\hskip 4pt}c@{\hskip 4pt}c@{\hskip 4pt}c@{\hskip 4pt}c@{\hskip 4pt}c@{\hskip 4pt}|>{\columncolor{gray!20}}c}
    \toprule
    Methods & Bus   & Bike  & Car   & Motor & Person & Rider & Truck & $mAP_{50}$ \\
    \midrule
    Source-FR\cite{Ren_He_Girshick_Sun_2017} &34.5     &29.6     &49.3      &26.2      &33.0       &35.1      &26.7     &33.5  \\
    SW\cite{Pan_Zhan_Shi_Tang_Luo_2019}    &30.6      &36.2       &44.6       & 25.1      &30.7      &34.6       &23.6       &30.8   \\
    IBN-Net\cite{Pan_Luo_Shi_Tang_2018} &29.9       &26.1       &44.5       &24.4       &26.2       &33.5       &22.4       &29.6   \\
    IterNorm\cite{Huang_Zhou_Zhu_Liu_Shao_2019} &29.7       &21.8       &42.4       & 24.4      &26.0       &33.3       & 21.6      &28.4   \\
    ISW\cite{Choi_Jung_Yun_Kim_Kim_Choo_2021}   &29.5       & 26.4      &49.2       &27.9       &30.7       &34.8       &24.0       &31.8   \\
    SDGOD\cite{wu2022single} &32.9       &28.0       &48.8       &29.8       &32.5       &38.2       &24.1       &33.5   \\
    CLIP-Gap\cite{vidit2023clip} &36.2       &34.2       &57.9       &\textbf{34.0}       &38.7       &43.8       &25.1       &38.5   \\
    DivAlign\cite{10656467}   & -      & -     &-       & -      & -     &-      &-     &37.2  \\
    UFR\cite{liu2024unbiased}   & 36.9      &\textbf{35.8}       &61.7       &33.7       & 39.5      &42.2       &27.5       &39.6   \\
    SRCD\cite{10742956}   & 36.4     &30.1     &52.4       &31.3       &33.4      &40.1       &27.7      &35.9   \\
    \midrule
    \midrule
    DINO\cite{zhang2022dino} &35.1       &30.1       &62.8       &29.0       &43.3       &41.3       &24.4       &38.0   \\
    SA-DETR & \textbf{38.0}      &31.6       &\textbf{67.1}  &31.7       &\textbf{48.3}       &\textbf{46.1}       &\textbf{28.9}       &\textbf{41.7}   \\
    \bottomrule
    \end{tabular}%
  }
  \label{tab:Daytime-Foggy}%
\end{table}%

\begin{table}[tp]
  \centering
  \caption{Experimental results (\%) on Night-Clear Scene.}
   \scalebox{1.0}{%
\begin{tabular}{c@{\hskip 4pt}|c@{\hskip 4pt}c@{\hskip 4pt}c@{\hskip 4pt}c@{\hskip 4pt}c@{\hskip 4pt}c@{\hskip 4pt}c@{\hskip 4pt}|>{\columncolor{gray!20}}c}
    \toprule
    Methods & Bus   & Bike  & Car   & Motor & Person & Rider & Truck & $mAP_{50}$ \\
    \midrule
    Source-FR\cite{Ren_He_Girshick_Sun_2017} &43.5     &31.2     &49.8      &17.5      &36.3       &29.2      &43.1     &35.8  \\
    SW\cite{Pan_Zhan_Shi_Tang_Luo_2019}    &38.7      &29.2       &49.8       &16.6      &31.5      &28.0       &40.2      &33.4   \\
    IBN-Net\cite{Pan_Luo_Shi_Tang_2018} &37.8       &27.3       &49.6       &15.1       &29.2       &27.1       &38.9       &32.1   \\
IterNorm\cite{Huang_Zhou_Zhu_Liu_Shao_2019} &38.5       &23.5       &38.9       &15.8       &26.6       &25.9       &38.1       &29.6   \\
    ISW\cite{Choi_Jung_Yun_Kim_Kim_Choo_2021}   &38.5       &28.5       &49.6       &15.4       &31.9       &27.5       &41.3       &33.2   \\
    SDGOD\cite{wu2022single} &40.6       &35.1       &50.7       &19.7       &34.7       &32.1       &43.4       &36.6   \\
    CLIP-Gap\cite{vidit2023clip} &37.7       &34.3       &58.0       &19.2       &37.6       &28.5       &42.9       &36.9   \\
    DivAlign\cite{10656467}   & -      & -     &-       & -      & -     &-      &-     &42.5  \\
    UFR\cite{liu2024unbiased}   &43.6       &38.1       &66.1       &14.7       &49.1       &26.4       &47.5       &40.8   \\
    SRCD\cite{10742956}   &43.1      &32.5     &52.3       & 20.1      &34.8      &31.5       &42.9      &36.7   \\
    \midrule
    \midrule
    DINO\cite{zhang2022dino} &44.1       &36.9       &71.0       &\textbf{20.2}       &54.8       &29.9       &49.0       &43.7   \\
    SA-DETR &\textbf{49.1}       &\textbf{41.1}       &\textbf{72.7}       &15.3       &\textbf{57.0}       &\textbf{33.8}       &\textbf{50.5}       &\textbf{45.6}   \\
    \bottomrule
    \end{tabular}%
  }
  \label{tab:Night-Clear}%
\end{table}%

\subsection{Visualization and Analysis} 

\subsubsection{Feature visualization} 
\begin{figure}[htbp]
\centering
\includegraphics[width=8cm]{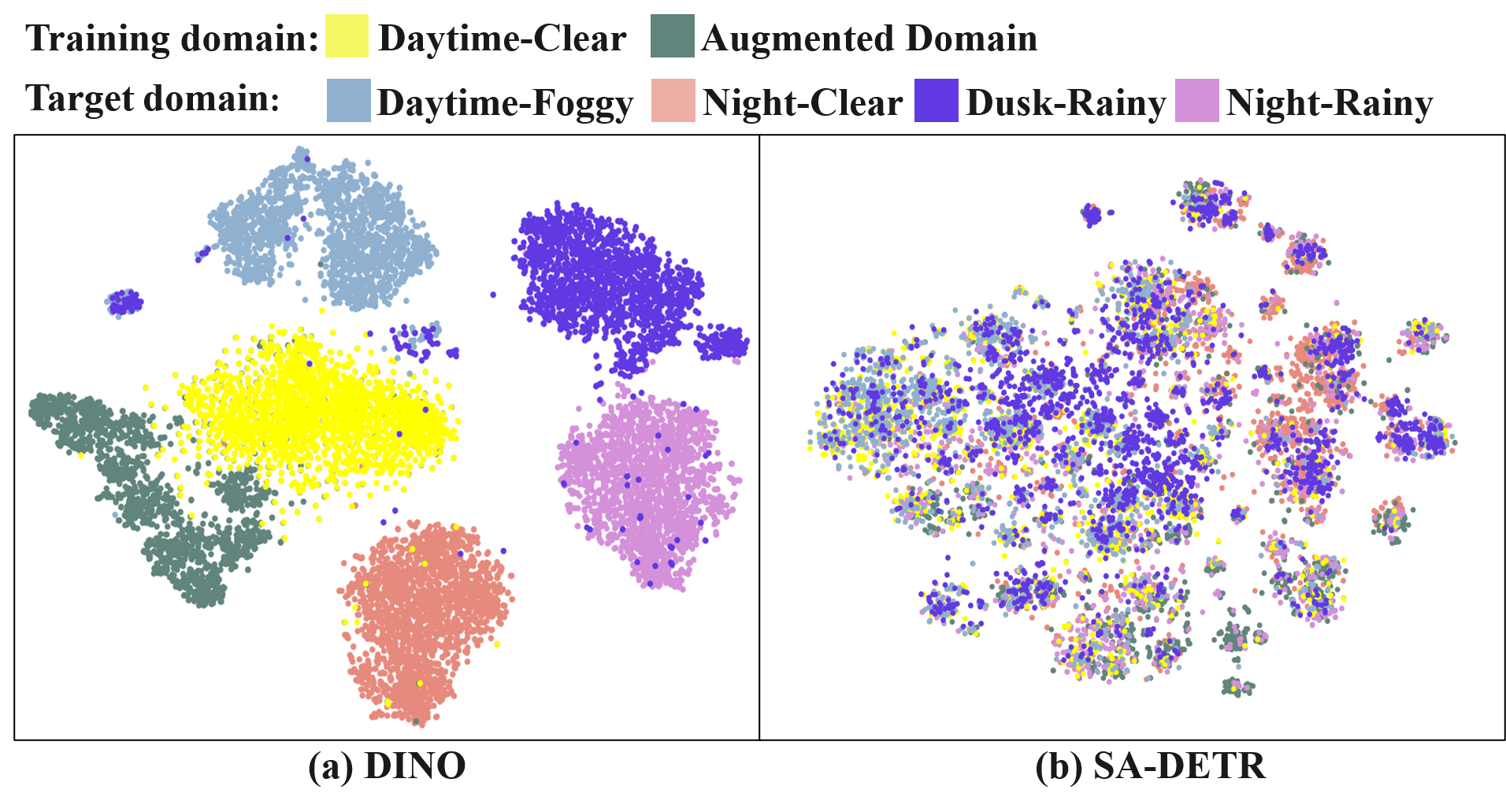}
\caption{The t-SNE visualization of style statistics between different domains, where the style statistics (concatenation of mean and variance) is computed from the last layer's feature map of the ResNet-50.}
\label{fig_3}
\end{figure}

We employ the t-distributed Stochastic Neighbor Embedding (t-SNE) method\cite{Maaten_Hinton_2008} to analyze the effectiveness of the proposed DSA and OCL module. First, we separately show the style distribution differences among various domains for the vanilla DINO and SA-DETR, as illustrated in Fig. \ref{fig_3}. It can be observed that the style distributions across various domains are well-separated in the vanilla DINO. Conversely, our ODS-Adapter projects the style representation of the target domain into the source domain, thereby entangling the style distributions across various domains. Through this dynamic adaptation to unseen scenarios, our adapter can more flexibly respond to diverse environmental changes, thereby effectively enhancing the detector's generalization capability.

\begin{figure}[htbp]
\centering
\includegraphics[width=8cm]{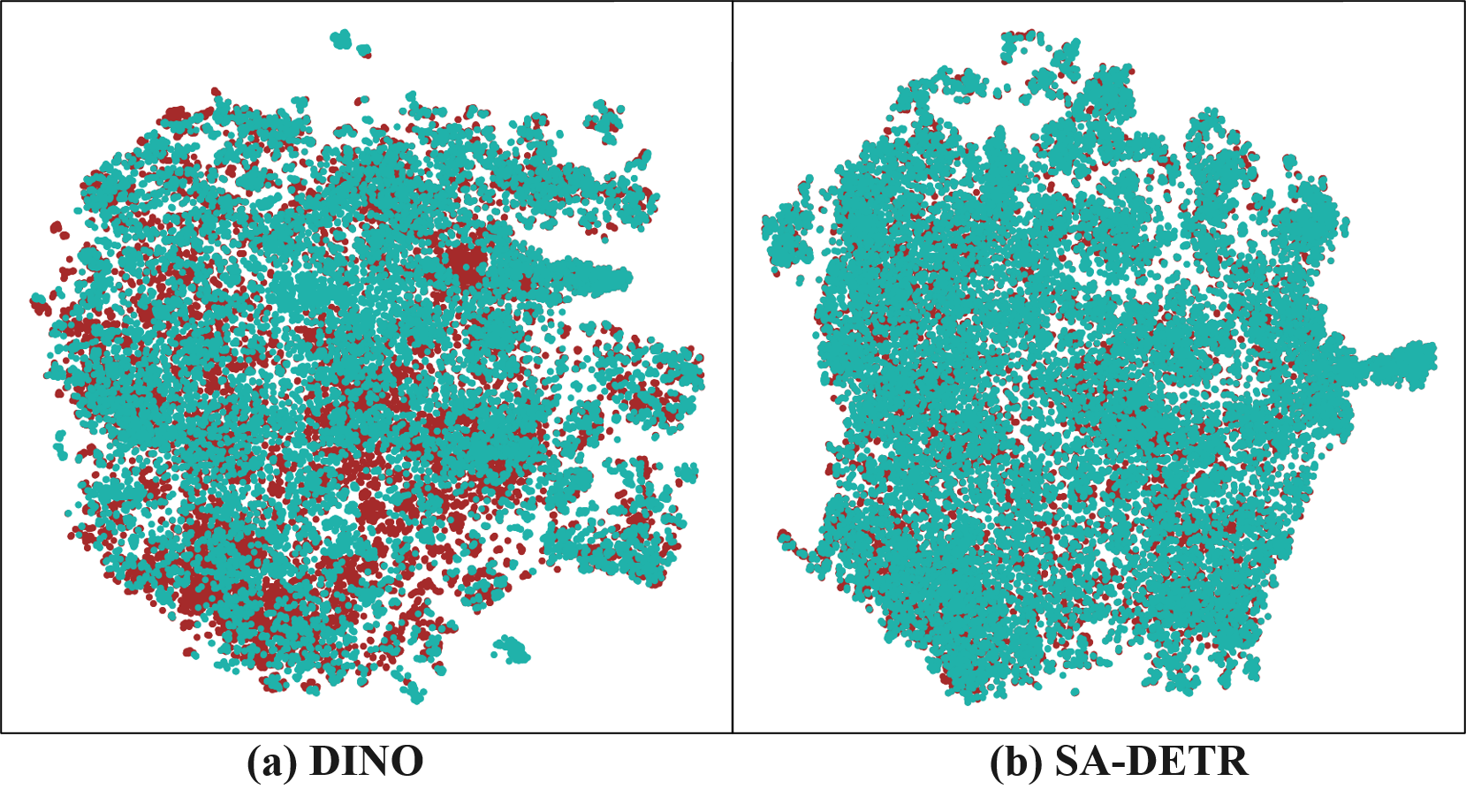}
\caption{The t-SNE visualization of object features from images originating from two training datasets, where red represents the source domain and green represents the augmented domain. Our OCL module aligns the domain gap well compared to the baseline method, thereby achieving predictions using instance-level domain-invariant features.}
\label{fig_4}
\end{figure}

Then, we visualized the features extracted from object queries, and Fig. \ref{fig_4} shows that our method exhibits a minimal domain gap compared to the baseline. This demonstrates that our OCL module successfully aligns features between the source and the augmented domains, thereby achieving predictions using domain-invariant instance-level features.

\begin{figure}[htbp]
\centering
\includegraphics[width=8cm]{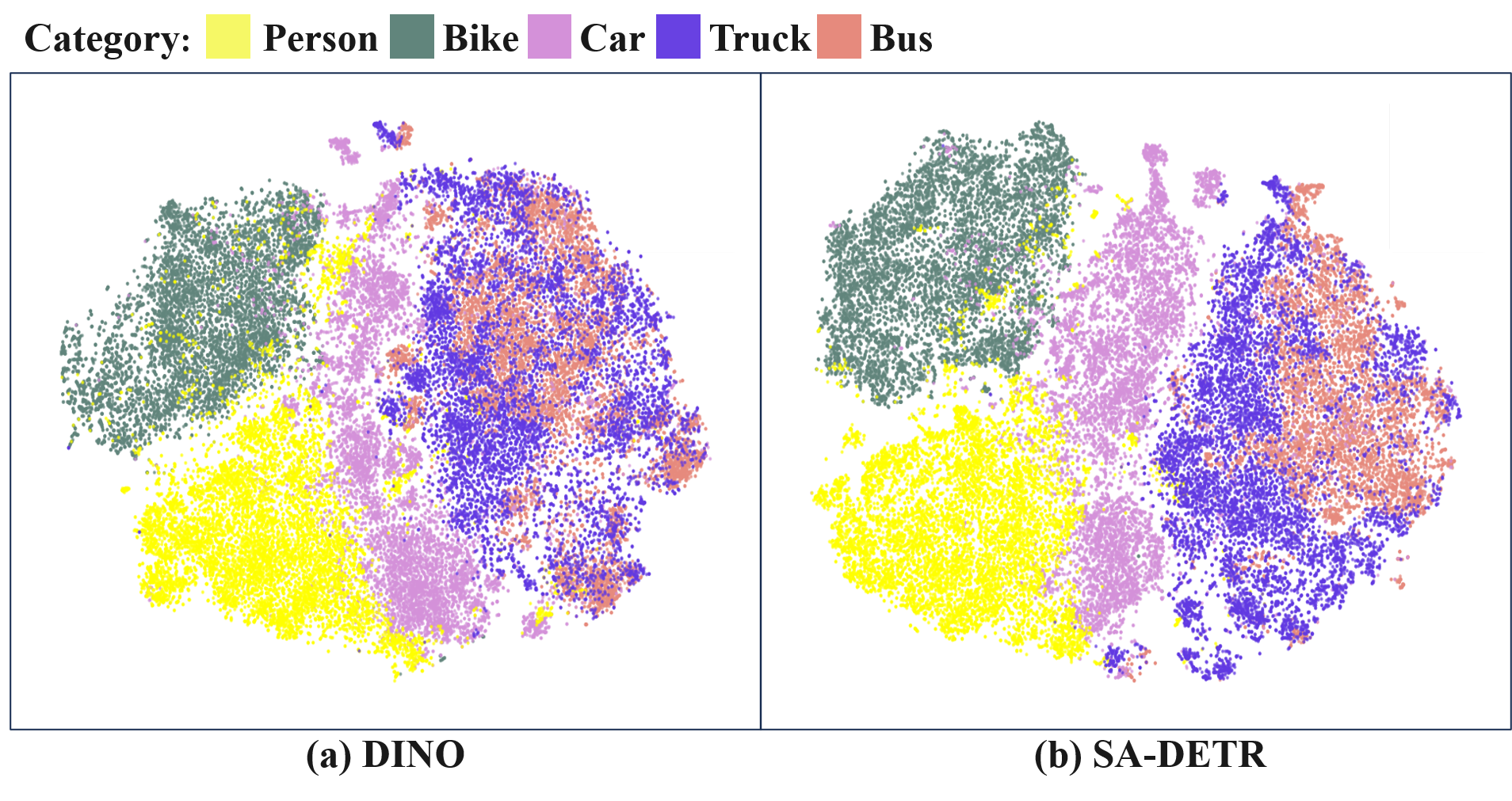}
\caption{The t-SNE visualization of object features from the five object categories in the Daytime-Clear images demonstrates that our detector benefits from contrastive learning, which enhances inter-category discriminability in the resultant feature space.}
\label{fig_5}
\end{figure}

Furthermore, we visualize object features by class category in the source domain, as shown in Fig. \ref{fig_5}. This demonstrates that our detector also benefits from contrastive learning. By attracting prototypes of the same category and repelling those of different categories across domains, this approach enhances the inter-class discriminability of our detector.

\begin{figure*}[htbp]
\centering
\includegraphics[width=18cm]{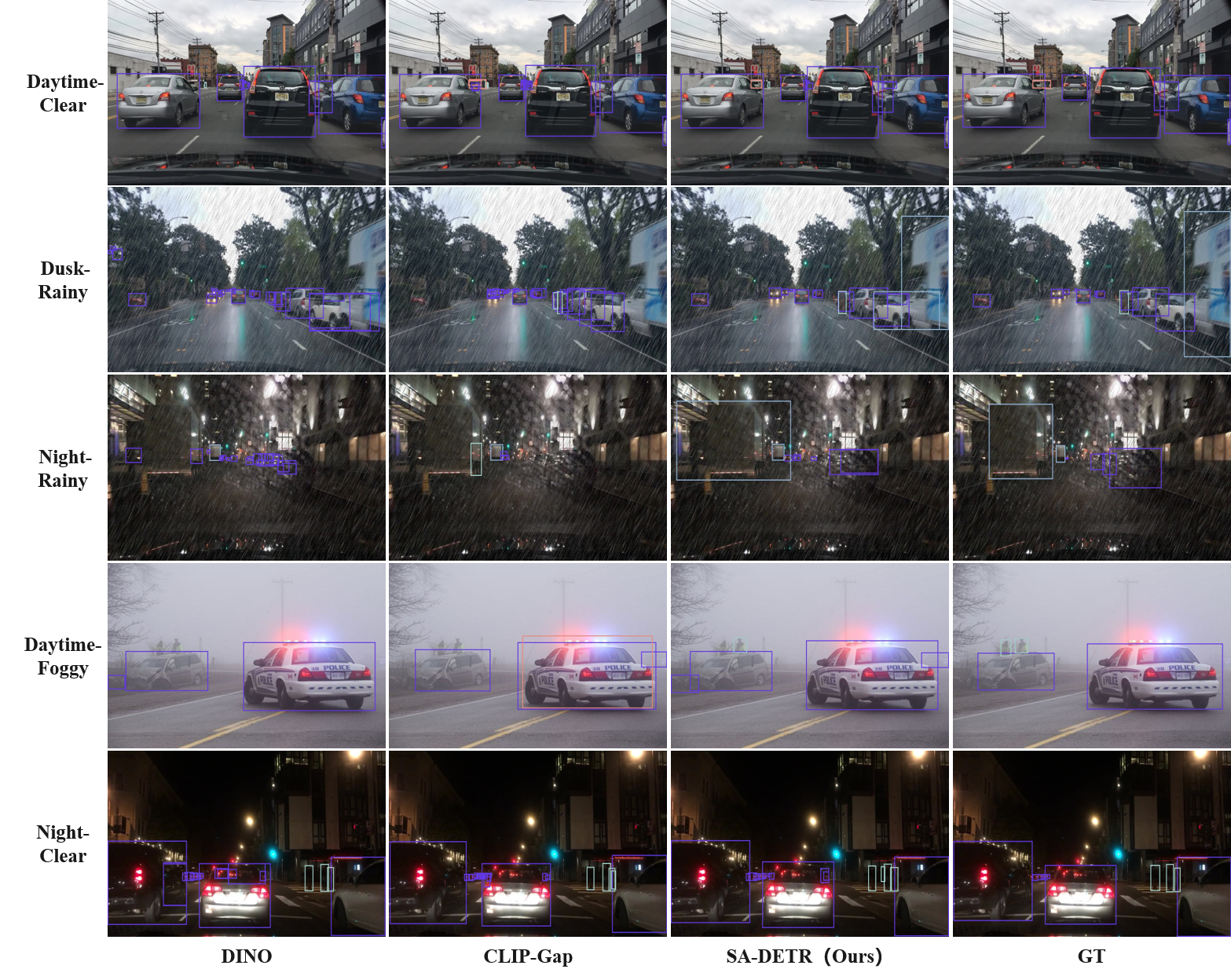}
\caption{Our detection results compared with the state-of-the-art methods. Different categories are marked with different colors. The confidence threshold for visualization is set to 0.2. "DINO" represents the base detector that uses only source domain data for training, and "GT" stands for Ground Truth. Other descriptions represent their respective methods.}
\label{fig_6}
\end{figure*}

\subsubsection{Detection results} We present the visualization results of SA-DETR across all scenarios, alongside those of vanilla DINO\cite{zhang2022dino}, Clip-gap\cite{vidit2023clip}, and Groundtruth, as shown in Fig. \ref{fig_6}. Compared to CNN-based methods, DETR-based detectors demonstrate greater accuracy in detecting small and weakly featured objects, which can be attributed to the transformer architecture’s effectiveness in capturing global structural information. Building on this, SA-DETR demonstrates more accurate classification results and higher recall rates when handling unseen scenarios, indicating that our proposed approach effectively enhances the detector's generalization capability. The visual results align with the numerical evaluations, indicating that our method exhibits excellent performance and generalization capability in both the source domain and various unseen target domains.

\section{Conclusion}

This paper introduces the Style-Adaptive DEtection TRansformer (SA-DETR), an effective and efficient DETR-based detector designed for single-source domain generalization (SDG) in object detection. To the best of our knowledge, this is the first work that explores a DETR-based detector for the SDG task. The SA-DETR adopts DINO as the base detector and incorporates an online domain style adapter and an object-aware contrastive learning module. Firstly, the adapter projects the style representation from the target domain onto the source domain, enabling dynamic style adaptation across various unseen scenarios. This adapter constructs a dynamic memory bank that self-organizes into diverse style bases and continuously absorbs statistics from unseen domains under a test-time adaptation framework, facilitating rapid and accurate style adaptation to previously unseen scenarios. Then, the proposed module uses annotations to constrain the scope of contrastive learning in both position and category, encouraging the detector to extract instance-level domain-invariant features to further enhance the detector's generalization capabilities. Extensive experiments across five distinct scenarios show that SA-DETR achieves superior performance for single-source domain generalization in object detection. In future work, we plan to explore methods for domain adaptation under conditions of small sample sizes.

\bibliographystyle{IEEEtran}
\bibliography{main}

\vfill

\end{document}